\documentclass{article}

\usepackage{arxiv}

\usepackage[utf8]{inputenc} 
\usepackage[T1]{fontenc}    
\usepackage{hyperref}       
\usepackage{url}            
\usepackage{booktabs}       
\usepackage{amsfonts}       
\usepackage{nicefrac}       
\usepackage{microtype}      
\usepackage{bm}             
\usepackage{amsmath}        
\usepackage{graphicx}       
\usepackage{caption}        
\usepackage{subcaption}     
\usepackage{mathtools}      
\usepackage{xcolor}         
\usepackage{multirow}       

\DeclarePairedDelimiter{\ceil}{\lceil}{\rceil} 
\DeclareMathOperator*{\argmax}{\text{argmax}}  
\DeclareMathOperator*{\argmin}{\text{argmin}}  
\DeclareMathOperator*{\trans}{\text{T}}  

\usepackage{algorithm}
\usepackage{algorithmicx}
\usepackage{algpseudocode}
\usepackage{tabularx}
\makeatletter
\newcommand{\multiline}[1]{%
  \begin{tabularx}{\dimexpr\linewidth-\ALG@thistlm}[t]{@{}X@{}}
    #1
  \end{tabularx}
}
\makeatother

\title{Edge Tracing using Gaussian Process Regression}

\author{
  Jamie Burke \\
  School of Mathematics\\
  James Clerk Maxwell Building\\
  The University of Edinburgh\\
  Edinburgh, UK EH9 3FD \\
  \texttt{James.Burke@ed.ac.uk} \\
   \And
  Stuart King \\
  School of Mathematics\\
  James Clerk Maxwell Building\\
  The University of Edinburgh\\
  Edinburgh, UK EH9 3FD \\
  \texttt{S.King@ed.ac.uk} \\
}

\begin{document}
\maketitle

\begin{abstract}\label{abstract}
We introduce a novel edge tracing algorithm using Gaussian process regression. Our edge-based segmentation algorithm models an edge of interest using Gaussian process regression and iteratively searches the image for edge pixels in a recursive Bayesian scheme. This procedure combines local edge information from the image gradient and global structural information from posterior curves, sampled from the model's posterior predictive distribution, to sequentially build and refine an observation set of edge pixels. This accumulation of pixels converges the distribution to the edge of interest. Hyperparameters can be tuned by the user at initialisation and optimised given the refined observation set. This tunable approach does not require any prior training and is not restricted to any particular type of imaging domain. Due to the model's uncertainty quantification, the algorithm is robust to artefacts and occlusions which degrade the quality and continuity of edges in images. Our approach also has the ability to efficiently trace edges in image sequences by using previous-image edge traces as a priori information for consecutive images. Various applications to medical imaging and satellite imaging are used to validate the technique and comparisons are made with two commonly used edge tracing algorithms.
\end{abstract}


\section{Introduction}
Semantic image segmentation is an important tool for deriving information from an image and plays a crucial role in many applications, facilitating higher-level image analysis \cite{song2019vision, zahoor2017fast, ulmas2020segmentation}. The goal of any image segmentation algorithm is to partition an image into meaningful sub-regions. Edge-based segmentation algorithms do this by identifying boundaries, or edges between those regions. Human visual systems are able to interpret and recognise scenes and objects just based on the basic structure of their outlines \cite{walther2011simple}, making edge-based approaches useful in mimicking some of the mechanisms of our own vision. Image segmentation is an active area of research since it is generally an ill-posed problem formulation. Consequently, much use is made of domain-specific implementations as no single segmentation algorithm suits all types of images.

This technique provides a tunable, robust, semi-automatic and probabilistic approach to edge-based image segmentation in images. The proposed methodology can be seen as a recursive Bayesian scheme, modelling the edge of interest (EoI) as a Gaussian process in a regression setting. Using a suitable edge map and tunable initialisation, the model iteratively searches for edge pixels, sequentially building an observation set to update and converge the posterior predictive distribution toward the EoI. This technique utilises the predictive variance of this distribution to search regions where no edge pixels have been observed and models uncertainty in regions with pixels already observed. This uncertainty quantification aids the tracing of edges which are degraded by noise and occlusions. There is no prior training required which allows the edge tracing technique to be applied to various different research fields. Moreover, the internal (hyper)parameters of the model allows the user to tune the fitting procedure to the EoI. For example, through specification of observation noise and choice of kernel, both of which have a strong impact on the fitting procedure.

\subsection{Related Work}\label{review}
Many early approaches to image segmentation used edge detection methods based on zero-crossings, discrete derivative operators or histogram thresholding \cite{prewitt1970object, duda1970experiments, roberts1963machine, rosenfeld1969picture}. A variety of these early approaches to image segmentation can be found in Rosenfeld's work \cite{rosenfeld1976digital}. The problem of edge detection has been well studied and with methods such as Canny \cite{canny1986computational}, it can be straightforward to obtain some form of edge map, or image gradient. However, there is a distinct difference between identifying the edges in an image and linking them to form a meaningful contour, with direct knowledge of the pixel coordinates. Szeliski \cite{szeliski2010computer} notes that this latter task is far more useful than the former and is more challenging due to the potential for extracting a poor quality edge map. This paper explicitly looks at solving this edge tracing task, with an edge map already obtained.

At present, the most popular frameworks employed to perform image segmentation are based on convolutional neural network models. These can produce highly accurate semantic or instance segmentation of images or videos \cite{ren2015faster, ronneberger2015u, milletari2016v}. For literature reviews on deep learning techniques for image segmentation see Wu et al. \cite{wu2019review} and for medical image segmentation see Hesamian et al. \cite{hesamian2019deep}. Such methods require large amounts of high quality, labelled training data to perform well on a given type of image problem and do not model the edge structure of an image directly. Moreover, Renard et al. \cite{renard2020variability} discuss the lack of interpretability and concern regarding the reproducibility of such methods in the medical image domain. Consequently, for many interesting applications it is still useful to work with edge-based segmentation models which are able to segment an image without training data and can provide a functional form of a given edge within an image.

Two common types of models suited for tracing individual edge structures are those which build a graph structure connecting the pixels within an image and those which adapt a contour to an edge structure using parametric energy-minimising splines. The latter are known as active contour models, originally introduced by Kass et al. \cite{kass1988snakes}. For extensive graph-based literature reviews for image segmentation see Peng et al. \cite{peng2013survey} and in medical image segmentation see Chen and Pan \cite{chen2018survey}. For detailed literature reviews on active contour models in image segmentation see Jaiswal and Sarode \cite{jaiswal2017review} and for medical image segmentation see Hemalatha et al. \cite{hemalatha2018active}.

The use of Bayesian frameworks in computer vision and image processing is popular due to such models being able to feature non-parametric learning and uncertainty quantification. Some image processing applications involve denoising \cite{liu2007using}, object categorisation \cite{kapoor2010gaussian}, detection of pneumonia in chest X-rays \cite{frank2020gaussian} or super-resolution of images \cite{he2011single}. Relevance vector machines \cite{tipping2001sparse} are related to Gaussian processes in its Bayesian formulation but has flexibility on kernel selection using localised basis functions and can induce sparsity due to its hyperparameter optimisation framework. Although this technique has been used in the computer vision community for classification \cite{dong2012accelerating, wei2005relevance}, there is a general consensus that sparsity and kernel flexibility comes at a cost of unnatural variance behaviour in the predictive distributions, unlike in Gaussian processes \cite{quinonero2004learning, rasmussen2005healing, martino2021joint, bishop2006pattern, rasmussen2003gaussian}, which can lead to undesirable results in modelling. We will not pursue relevance vector machines in this paper because the proposed edge tracing methodology relies heavily on the predictive variance for image exploration. Relevance vector machines do not have the capacity to trace well because of this unnatural variance behaviour.

There has not been an extensive amount of work on applying Gaussian processes to image segmentation. Freytag et al. \cite{freytag2012efficient} perform semantic segmentation by using histogram intersection kernels for fast and exact Gaussian process classification. Tu and Zhu \cite{tu2002image} proposed a Bayesian statistical framework to image segmentation by learning object appearance models for different region types, placing prior distributions over region size, region number and boundary smoothness. Simek and Barnard \cite{simek2015gaussian} used a two-dimensional Gaussian process regression model to segment Arabidopsis leaves by modelling the blade and petiole as two random functions, joining them at their boundaries using a smoothing constraint. Our approach is in a similar vein to Simek and Barnard, but there is no trained prior based on a priori information. 

\section{Gaussian Process Regression}\label{gpr}
In Bayesian machine learning, a Gaussian process regressor (GPR) is viewed as a multivariate Gaussian distribution over function values. They are primarily used to model complex functions and quantify uncertainty in model fitting to continuous, real-valued data. The GPR is able to treat the learning process incrementally by updating beliefs in regions where observations can be incorporated. This belief, or uncertainty, is induced by the prior predictive distribution and influenced by subsequent observations which give rise to the posterior predictive distribution (PPD). The GPR models the underlying function $f(x)$ as a Gaussian process using noisy observations $y$ such that
\begin{equation}\label{eq:gpr_assume_model}
    y = f(x) + \epsilon,
\end{equation}
with $f(x)$ fully specified by its mean function $\mu(x)$ and kernel $k(x, x')$ defined by
\begin{align}
    \mu(x) &= \mathbb{E}\left[f(x)\right];\\
    k(x, x') &= \mathbb{E}\left[(f(x)-\mu(x))(f(x')-\mu(x'))\right];\\
    f(x) &\sim \mathcal{GP}\big(\mu(x), k(x, x')\big)\label{eq:gp_defn}.
\end{align}
For continuity with the remainder of this paper, we have assumed a one-dimensional domain and range, $x, x', f(x) \in \mathbb{R}$, as the proposed methodology works with a 2D image, utilising the real-valued intervals on the horizontal and vertical axes as the domain and range of the GPR. The identically, independently distributed Gaussian noise $\epsilon \sim \mathcal{N}\big(0, \sigma_y^2\big)$ is characterised by the observation noise variance $\sigma_y^2$ which governs the uncertainty that the model assumes in the observation set. Without any loss of generality we can assume a trivial mean function, but it is the choice of kernel which has the largest impact on model fitting. The kernel is used to build the covariance matrix of random function values, which depend on their input observations. 

In this work we are only concerned with isotropic, stationary kernels which are translation invariant, monotonic and dependent on the distance between input locations, $r_{ij} = | x_i - x_j |$. The kernel formulation imposes differentiability constraints while its hyperparameters dictate amplitude and sinuosity of the random curves which can be sampled from the PPD. It is common for these kernels to have at least two hyperparameters, the signal variance, $\sigma_f^2$ and lengthscale, $\ell$. $\sigma_f^2$ dictates the average amplitude, or deviation, of the random curves from the mean while $\ell$ roughly describes the typical distance between turning points, effectively controlling their sinuosity.

The squared exponential and Mat\'{e}rn kernels are the isotropic, stationary kernels of interest in this paper. These kernels are scaled by a Gaussian kernel allowing them to compute covariance between function values monotonically with respect to the distance between their corresponding input locations, $k = k(r_{ij};\hspace{2pt} \sigma_f^2, \ell)$. Although the squared exponential kernel is commonly used for drawing infinitely differentiable random curves, the strong assumption on continuity is one of its pitfalls when modelling real-life physical processes, as these tend to be more irregular. 

Consequently, the Mat\'{e}rn class of kernels have more flexibility in the degree of differentiability and are preferred for modelling processes which are less smooth. They are enumerated by their unique hyperparameter $\nu$, dictating the degree of differentiability of random curves they can generate. These kernels have been observed by Rasmussen and Williams \cite{rasmussen2003gaussian} to be particularly useful when $\nu = 3/2$ or $\nu=5/2$. These values yield first-order and second-order differentiable curves from the PPD, respectively. This makes this kernel an ideal choice for modelling edges in real-life image segmentation problems. More information on these kernels, their theory and formulation can be found in Rasmussen and Williams' book on Gaussian processes for machine learning \cite{rasmussen2003gaussian}.

\subsection{Posterior Predictive Distribution}
After selection of a valid kernel $k$ and its kernel hyperparameters, we can construct the prior predictive distribution, from which we can draw random prior curves, $\bm{f}^*$ evaluated along some domain $X^*$ such that
\begin{equation}
    \bm{f}^* \sim \mathcal{N}\big(\bm{0}, K_{**}\big),
\end{equation}
with $K_{**}=K\left(X^*, X^*\right) = \big[k(x_i, x_j)\big]_{ij}$. However, in model fitting, we are more interested in using an observation set that has the form of equation \eqref{eq:gpr_assume_model} to model an underlying function or process. These noisy observations, $\mathcal{D} = \big\{X, \bm{y}\big\}$ are integrated into the GPR by forming the joint distribution between them and the function values we seek but are unknown, $\bm{f}^*$. The covariance matrix of this distribution is a block matrix comprising of the covariance pairs between the known target values, $\bm{y}$ and the unknown function values, $\bm{f}^*$ using their corresponding input locations, $X$ and $X^*$, respectively. This joint distribution is defined by
\begin{equation}\label{eq:jointdist}
    \begin{bmatrix} \bm{y} \\ \bm{f}^*\end{bmatrix} \sim \mathcal{N}\bigg(\bm{0}, \begin{bmatrix} K + \sigma_y^2\mathbb{I} & K_* \\ K_*^{\trans} & K_{**}\end{bmatrix}\bigg),
\end{equation}
where $K_*=K(X, X^*)=K(X^*, X)^{\text{T}}$. Here, we encode the observation noise into the covariance matrix between observations, $\text{Cov}[\bm{y}] = K(X, X) + \sigma_y^2\mathbb{I} = K +\sigma_y^2\mathbb{I}$. This observation noise promotes numerical stability and quantifies the freedom posterior curves have in not intersecting observations exactly. Collating all kernel hyperparameters and observation noise into $\bm{\theta}$, these are parameters which can be fixed during model fitting or optimised through maximising the log marginal likelihood of the observation set,
\begin{align}
    \bm{\theta} &= \argmin_{\bm{\theta}}\Big\{\text{log}\hspace{3pt} p(\bm{y} \hspace{3pt} | \hspace{3pt} X, \bm{\theta})\Big\};\label{eq:optim_theta}\\ \label{eq:marginal_likelihood}
    \text{log} \hspace{3pt} p(\bm{y} \hspace{3pt} | \hspace{3pt} X, \bm{\theta}) &= -\frac{1}{2}\bm{y}^{\trans}A^{-1}\bm{y} - \frac{1}{2}\text{log}\hspace{1pt}\big|A\big| - \frac{m}{2}\text{log}\hspace{1pt}2\pi,
\end{align}
with $m$ the sample size and $A = K+\sigma_y^2\mathbb{I}$. Once $\mathcal{D}$ and $\bm{\theta}$ are known we can restrict the joint distribution in equation \eqref{eq:jointdist} to only describe functions which accommodate the observations in $\mathcal{D}$, inducing the PPD. This is done by conditioning on this joint distribution using $\mathcal{D}$, $\bm{\theta}$ and the input locations of unknown function values $X^*$,

\begin{align}\label{eq:gp_posterior}
    \bm{f}^* \hspace{3pt}\big|\hspace{3pt} \mathcal{D} &\sim \mathcal{N}\big(\mathbb{E}\left[\bm{f}^*\hspace{3pt} | \hspace{3pt} X, \bm{y}\right], \text{Cov}\left[\bm{f}^* \hspace{3pt} | \hspace{3pt} X\right]\big);\\\label{eq:gp_updated_mean}
    \mathbb{E}\left[\bm{f}^* \hspace{3pt} | \hspace{3pt} X, \bm{y}\right] &= K_*^{\text{T}}\left[K+ \sigma^2_y\mathbb{I}\right]^{-1}\bm{y};\\ \label{eq:gp_updated_cov}
    \text{Cov}\left[\bm{f}^* \hspace{3pt} | \hspace{3pt} X\right] &= K_{**} - K_*^{\text{T}}\left[K+ \sigma^2_y\mathbb{I}\right]^{-1}K_*,
\end{align}
allowing us to sample posterior curves in a similar fashion to the prior predictive distribution. For the equations above and for the remainder of this paper, we implicitly assume conditioning on $\bm{\theta}$ and test input locations $X^*$ for readability. The mathematical details of constructing this PPD is shown in appendix A.2 of Rasmussen and Williams' book \cite{rasmussen2003gaussian}. 

\section{Methodology}\label{method}
The proposed procedure of tracing an EoI in an image of size $M \times N$ is as follows. In the first instance, a suitable choice of kernel $k$ and hyperparameters $\bm{\theta}$ are selected and fixed to model the EoI (more on this in section \ref{discussion}). An initial PPD, $\bm{f}^{(1)} | \mathcal{D}^{(0)}$, is induced through fitting an estimate for the edge endpoints, $\mathcal{D}^{(0)} = \big\{X^{(0)}, \bm{y}^{(0)}\big\}$. In each iteration $n$, $L$ posterior curves are sampled using the previous iteration's PPD, $\bm{f}^{(n)} | \mathcal{D}^{(n-1)}$, predicted at inputs along the image's horizontal axis, $X^* = [0, N-1] \subset \mathbb{Z}$. These curves are scored using the pre-computed image gradient $G$ and a proportion of the highest scoring posterior curves $\epsilon L$ are used to decide on the next set of pixels that will be fitted to the model's observation set $\mathcal{D}^{(n)}$ in the next iteration.

The individual elements of these optimal posterior curves are used to estimate a two-dimensional weighted smoothing kernel $\phi^{(n)}$ which represents a frequency distribution of where these optimal curves pass through the image. Combining this density function with the image gradient, pixels are scored and only the higher scoring pixels are chosen, according to an adaptive threshold $T$, to form the next observation set, $\mathcal{D}^{(n+1)} = \big\{X^{(n+1)}, \bm{y}^{(n+1)}\big\}$. Each consecutive iteration refits the Gaussian process with these new observations, updating the belief in the PPD on where the EoI lies. In the final iteration, with enough observations to reconstruct the EoI, the kernel hyperparameters $\bm{\theta}$ are optimised according to equation \eqref{eq:optim_theta} and the posterior predictive mean, $\mathbb{E}\left[\bm{f}^{(n)} \hspace{3pt} | \hspace{3pt} \mathcal{D}^{(n)}\right]$, evaluated at $X^*$, and a 95\% credible band is outputted to the user. While in sections \ref{apps} and \ref{discussion} we discuss circumventions for edges which are non-injective along the image width, for definiteness and simplicity we will assume the EoI can be modelled as an injective function and that its horizontal pixel length spans the width of the image, $N$. 

\subsection{Initialisation}
At the beginning of the procedure, following a similar formulation to equation \eqref{eq:gp_defn}, our prior distribution comes from a zero mean Gaussian process with kernel $k$ such that 
\begin{equation}
    f^{(0)}(x) \sim \mathcal{GP}\big(0, k(x, x' ; \bm{\theta})\big),
\end{equation}
inducing a prior predictive distribution where a vector of random function values $\bm{f}^{(0)}$ can be sampled using
\begin{equation}\label{eq:init_prior}
    \bm{f}^{(0)} \sim \mathcal{N}\big(\bm{0}, K_{**}\big).
\end{equation}
To induce the initial PPD requires an estimate for the edge endpoints, $\mathcal{D}^{(0)} = \big\{X^{(0)}, \bm{y}^{(0)}\big\}$. These user-specified pixel coordinates act as domain knowledge and are assumed to be within the vicinity of the true edge endpoints. As we may not know the true edge endpoints, these are fitted and imputed along with the remainder of the interior edge pixels. This is done by specifying the variance of the observation noise at these edge endpoints as $\sigma_y^2$. In practice, the estimate for the true edge endpoints could be known and the algorithm has the flexibility of assuming a non-uniform noise model by setting the observation noise variance for these pixels as a very low value for numerical stability. This initialisation step tells the algorithm which edge the user is interested in tracing. For the remainder of this section, we will assume a uniform noise model. Fitting $\mathcal{D}^{(0)}$ induces the initial PPD, $\bm{f}^{(1)} | \mathcal{D}^{(0)}$, from which we can draw samples evaluated at $X^*$.

\begin{figure}[!t]
    \centering
    \includegraphics[width=0.75\textwidth]{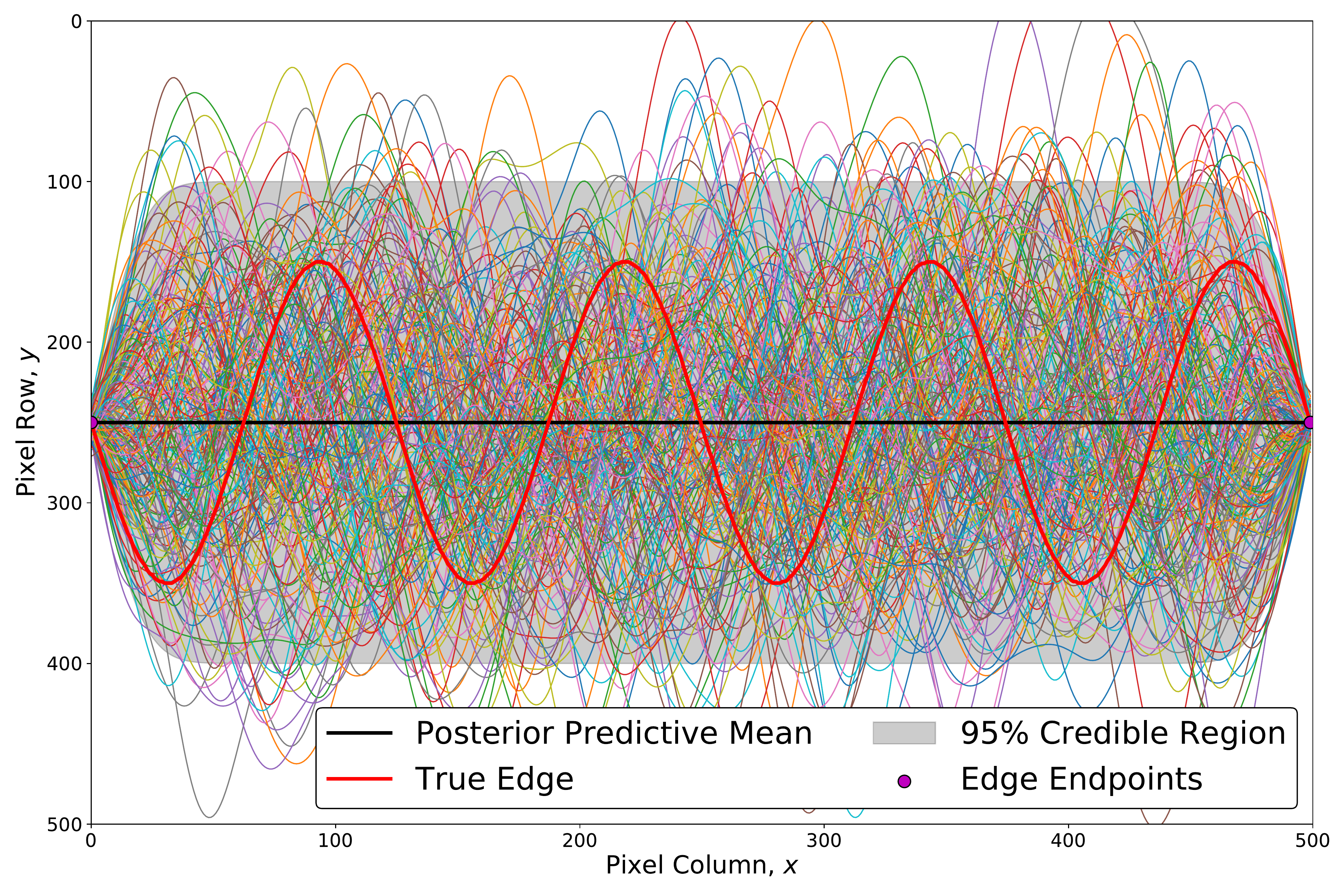}
    \caption{200 posterior curves drawn from an initial PPD. The grey shaded region represents the 95\% credible region. The squared exponential was chosen as the kernel with $\sigma^2_f = 75^2$, $\ell = 20$ and observation noise variance $\sigma^2_y = 1^2$.}
    \label{fig:init_post_dist}
\end{figure}

The objective is to now impute the missing pixels between the edge endpoints. The predictive variance of the PPD increases in regions far away from the edge endpoints, allowing the posterior curves to evolve freely across the image, according to the properties imposed by the selected kernel. In fact, the signal variance $\sigma_f^2$ dictates the maximum size of this predictive variance. With enough distance from the edge endpoints, the initial PPD will have a 95\% credible interval at maximum of $\overline{\bm{y}}^{(0)} \pm 2\sigma_f$, where $\overline{\bm{y}}^{(0)}$ is the mean of the edge endpoint pixel heights. This predictive variance is something we take advantage of throughout the fitting procedure so that posterior curves explore around the EoI and propose possible locations for it. Figure \ref{fig:init_post_dist} shows an initial PPD for an EoI, shown in bold red. The remainder of this section describes the methodology using an arbitrary iteration $n \geq 0$.

\subsection{Curve Scoring}\label{subsec:curve_scoring}
In each iteration, $L$ posterior curves $\bm{f}^{(n)}_l$, $l=1, \dots, L$ are sampled from the PPD, evaluated at $X^*$. These curves represent a sample of the model's belief in where the EoI lies, given the current observation set. To select which pixel coordinates to add to the observation set, these curves must be scored by how well they fit the EoI. This is done using the pre-computed and user-defined image gradient $G$. The score for a curve is computed as the cumulative value of the gradient response along this curve per unit pixel length, 
\begin{equation}
    I(\bm{f}^{(n)}_l) = \int_{\bm{f}^{(n)}_l} G\big(\bm{r}(s)\big) \hspace{2pt} ds \bigg/ \int_{\bm{f}^{(n)}_l} ds.\label{eq:LI_1}
\end{equation}

This score is analogous to computing the area under the posterior curve along the interpolated image gradient surface, divided by the posterior curve's arc length. We use Simpson's rule and finite difference rules to estimate the line integral and arc length. The advantage of scaling curve scores by arc length is that curves will be penalised for making longer excursions. This can be helpful in the presence of noise in the image gradient. The cost function for scoring a posterior curve is the inverse of this gradient score, $C(\bm{f}^{(n)}_l) = \big(I(\bm{f}^{(n)}_l)\big)^{-1}$. This is so that higher scoring curves are minimally costly and the minimisation of this cost becomes part of the optimisation scheme to model the edge. 

\subsection{Pixel Scoring}
An integer proportion $\epsilon L$, $\epsilon \in (0, 1]$, of posterior curves which have the lowest costs are selected as the most optimal posterior curves. These optimal curves provide a best guess at where the current PPD believes the EoI lies and their coordinates together map out which regions have been visited the most. Given these curves lie in the vicinity of the EoI, collating their coordinates to construct a weighted smoothing kernel of where they explore in the image will provide an empirical approximation for where the PPD believes the EoI is. This weighted, frequency density function is defined by  

\begin{equation}\label{eq:freq_kde}
    \phi^{(n)}(\bm{z}) = \frac{1}{\sum_{l, i}w(\bm{z}_{li})}\sum_{l, i}w(\bm{z}_{li})K_{\mathbb{I}}(\bm{z} - \bm{z}_{li}),
\end{equation}
where $\bm{z}, \bm{z}_{li} \in [0, N-1] \times [0, M-1] \subset \mathbb{R}^2$. $\bm{z}_{li}$ corresponds to the $i^{\textrm{th}}$ coordinate of the $l^{\textrm{th}}$ posterior curve, $\bm{f}^{(n)}_l$. The weighting for $\bm{z}_{li}$ scores the point according to how optimal the curve $\bm{f}^{(n)}_l$ is. This weighting will result in a higher density for regions where the most optimal curves lie in the image. The weight function $w$ is defined by
\begin{equation}\label{eq:kde_weight}
    w(\bm{z}_{li}) = \frac{I\big(\bm{f}^{(n)}_l\big)}{\sum^{\epsilon L}_{l=1}I\big(\bm{f}^{(n)}_l\big)}, \hspace{2pt} \forall i.
\end{equation}

$K_{\mathbb{I}}(\bm{z}-\bm{z}_{li})$ is the symmetric, multivariate kernel of the density function. Here we use the isotropic, two-dimensional Gaussian kernel with the identity matrix as its two-dimensional lengthscale,
\begin{equation}\label{eq:gauss_kernel}
    K_{\mathbb{I}}(\bm{z} - \bm{z}_{li}) = (2\pi)^{-1}\exp\left(-\frac{1}{2}\big(\bm{z} - \bm{z}_{li}\big)^{\text{T}}\big(\bm{z} - \bm{z}_{li}\big)\right).
\end{equation}

The isotropic and unitary lengthscale imposes no dominant orientation in the Gaussian kernel centred at each real-valued point $\bm{z}_{li}$. Moreover, the unitary property ensures each point's kernel contributes the majority of its density to only the 4 nearest pixel coordinates in the discretised space. 

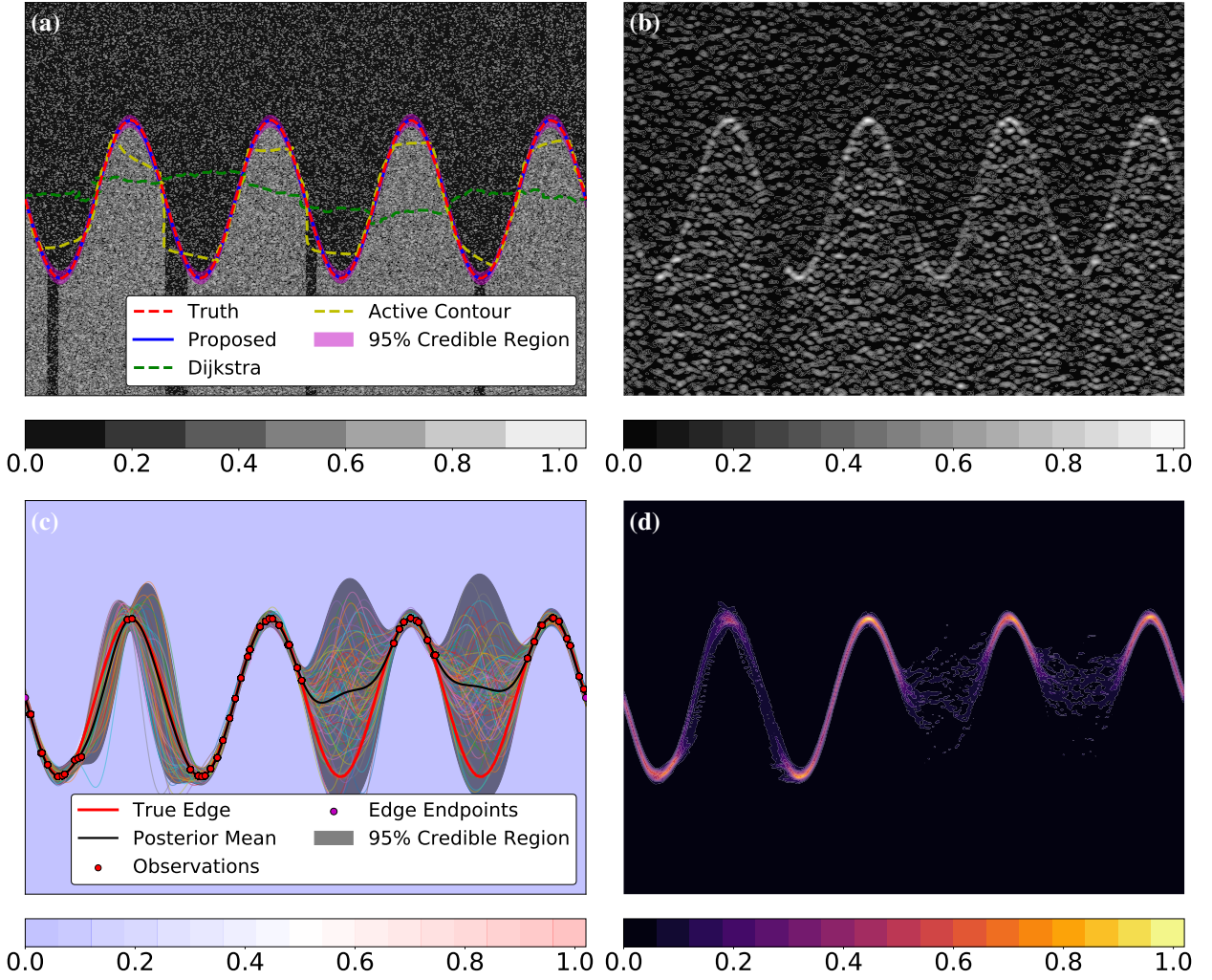
\begin{figure}[!t]
    \centering
    \setlength{\unitlength}{1cm}
    \begin{picture}(18,13.6)
        \put(0.0,0.0){\includegraphics[width=\textwidth]{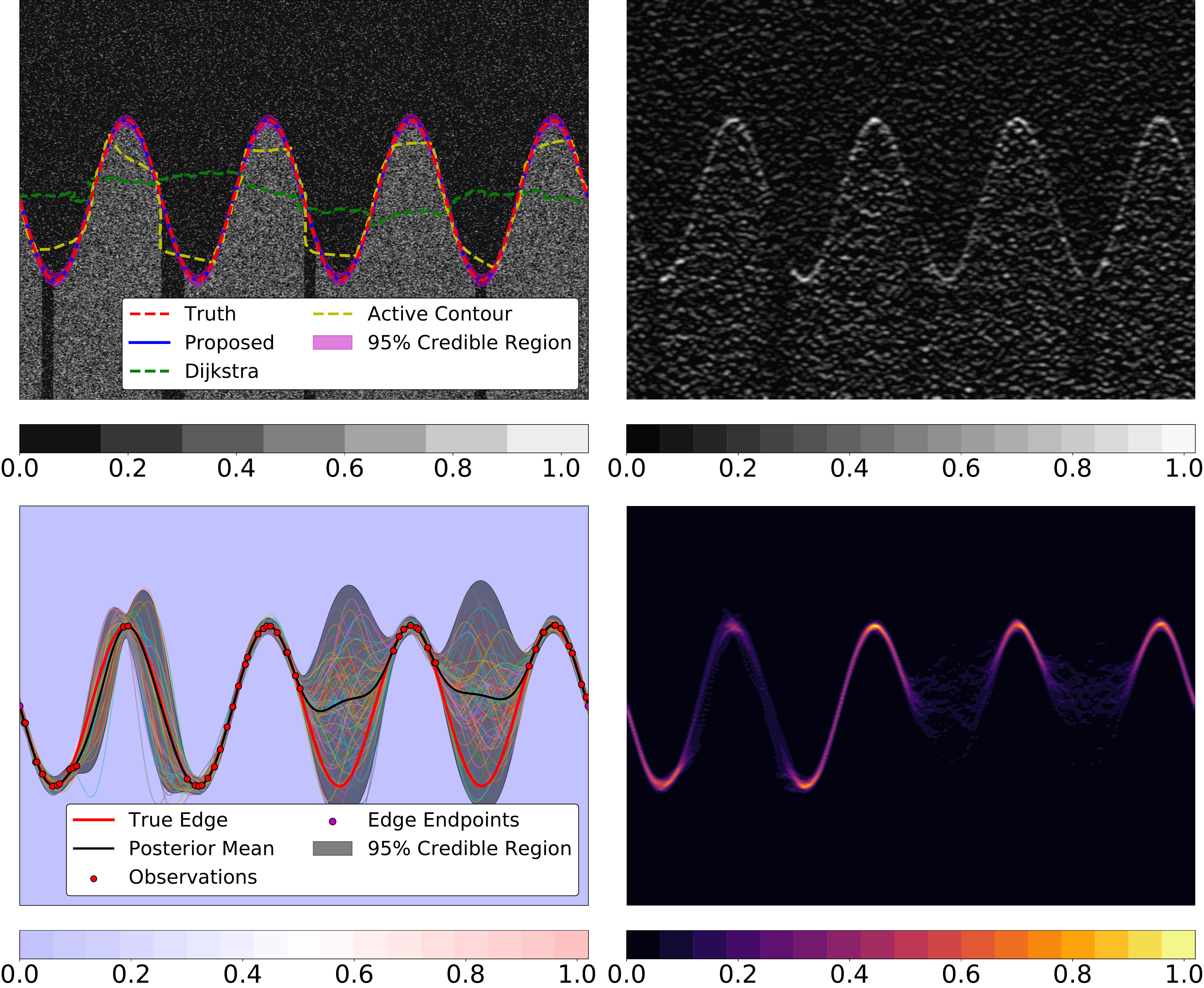}}
        \put(0.35, 13.15){\textbf{\color{white}(a)}}
        \put(8.675, 13.15){\textbf{\color{white}(b)}}
        \put(0.35, 6.2){\textbf{\color{white}(c)}}
        \put(8.675, 6.2){\textbf{\color{white}(d)}}
    \end{picture}
    \caption{(a) Noisy and occluded test case. (b) Image gradient. (c) $\epsilon L$ optimal posterior curves from a PPD fitted with an incomplete observation set. (d) Weighted frequency density function using optimal posterior curves from (c).}
    \label{fig:kde_plot}
\end{figure}

Figure \ref{fig:kde_plot} illustrates the process during the fitting procedure for a test case whose initial PPD is shown in figure \ref{fig:init_post_dist}. Subfigure \ref{fig:kde_plot}(a) shows a noisy and occluded sinusoidal edge. The true edge and edge predictions using the proposed methodology and two common edge tracing techniques are superimposed. A 95\% credible band is shown in magenta around the proposed edge prediction. Subfigure \ref{fig:kde_plot}(c) shows the $\epsilon L$ optimal posterior curves from the iteration's PPD with a 95\% credible interval represented by the grey shaded region. Note that almost all of the first two periods of the EoI are well mapped by the posterior curves, with a reduced 95\% credible interval in the region of dense observations. Conversely, the posterior curves are more variable in the latter two periods due to a lack of observations. These two remarks can be observed in subfigure \ref{fig:kde_plot}(d), where we see a significantly higher density covering the former two periods of the true edge in contrast to the latter.

Individual pixel coordinates, $\bm{p} = (x, y)$, can then be scored by combining their density, $\bm{\phi}^{(n)}(\bm{p})$ and gradient response, $G(\bm{p})$. This score, $s(\bm{p}) \in [0, 1]$, is defined by
\begin{equation}\label{eq:score_eqn}
    s(\bm{p}) = \frac{1}{3}\left(\phi^{(n)}(\bm{p}) \cdot G(\bm{p}) + \phi^{(n)}(\bm{p}) + G(\bm{p})\right).
\end{equation}

\subsection{Accept-Discard Scheme}
The primary advantage of using a score function which includes the iteration-dependent frequency density $\phi^{(n)}$ is that densities are updated in each consecutive iteration using the new PPD. This means that pixels fitted during previous iterations are re-scored using an updated density allowing a more accurate representation of how probable those pixels are of belonging to the EoI, given the new set of optimal posterior curves. This allows the possibility of older, incorrectly fitted pixels to be discarded from the model if their score drops sufficiently in later iterations. An adaptive, user-specified value $T \in [0, 1]$ is chosen to threshold these pixels such that
\begin{equation}\label{eq:thresh_set}
   P^{(n)}_T = \Big\{\bm{p} \hspace{3pt} \Big| \hspace{3pt} s(\bm{p}) \geq T\Big\}.
\end{equation}

The thresholded pixels are divided into $\ceil{\nicefrac{N}{\Delta x}}$ linearly spaced sub-intervals, $S_i = \left[(i-1)\Delta x, \hspace{3pt} i\Delta x\right]$, $i = 1, \dots, \ceil{\nicefrac{N}{\Delta x}}$. The length of each sub-interval is a user-specified parameter, $\Delta x$. As a continuous function is fitted, injectivity is enforced by using non-max suppression, selecting the highest scoring pixel per sub-interval, yielding
\begin{equation}\label{eq:accepted_set} 
    P^{(n)}_A = \Big\{\displaystyle\argmax_{\bm{p}_j}\big\{s(\bm{p}_j)\big\}\hspace{3pt} \Big| \hspace{3pt} x_j \in S_i\Big\}_{i = 1}^{\ceil{\nicefrac{N}{\Delta x}}}.
\end{equation}

Coordinates are binned because it is unnecessary to fit a GPR with points in very close proximity in the domain, unless the underlying edge is very complex. Because $\phi^{(n)}$ changes with each iteration, each sub-interval's highest scoring pixel coordinate may change dependant on the score it obtains compared to other possible pixels in its neighbourhood. The observation set for the next iteration $\mathcal{D}^{(n)}$ is formed by re-scoring the previous iteration's observation set according to the new set of optimal posterior curves, $\mathcal{D}_*^{(n-1)}$, and combining this with $P^{(n)}_A$ through binning and non-max suppression, yielding 
\begin{equation}\label{eq:concat_prev_new_set}
    \mathcal{D}^{(n)} = \textrm{suppression}\left(\textrm{binning}\big(P^{(n)}_A \cup \mathcal{D}_*^{(n-1)}\big)\right).
\end{equation}

\subsection{Refitting \& Convergence}
The observations in $\mathcal{D}^{(n)}$ represent the most probable pixel coordinates which form part of the EoI for the $n^{\text{th}}$ iteration. An improved PPD can be formed using this new set of pixel coordinates, updating the model's belief in where the EoI lies,
\begin{align}\label{eq:iter_posterior}
    \bm{f}^{(n+1)} \hspace{3pt}\big|\hspace{3pt}\mathcal{D}^{(n)} &\sim \mathcal{N}\big(\bm{\mu}^{(n+1)}, \bm{\sigma}^{(n+1)}\big);\\ \label{eq:iter_predictive_mean}
    \bm{\mu}^{(n+1)} &= K_{*n}\left[K_{nn}+ \sigma^2_y\mathbb{I} \right]^{-1}\bm{y}^{(n)};\\\label{eq:iter_predictive_cov} 
    \bm{\sigma}^{(n+1)} &= K_{**} - K_{*n}\left[K_{nn}+ \sigma^2_y\mathbb{I}\right]^{-1}K_{n*},
\end{align}
where $K_{nn} = K(X^{(n)}, X^{(n)})$ and similarly for $K_{n*}$ and $K_{*n}$. The posterior curves $\bm{f}^{(n+1)}$ we can sample are now more accurate in mapping regions of the EoI than in previous iterations. The user-specified observation noise variance $\sigma^2_y$ is used to perturb the posterior curves from passing through the observation points exactly, allowing posterior curves to continue searching for pixels in the image that are in the vicinity of previously fitted pixels. This permits the possibility of accepting better edge pixels in each sub-interval in future iterations.

The algorithm reaches termination once all sub-intervals have selected a high-scoring pixel, i.e. $|\mathcal{D}^{(n)}| = \ceil{\nicefrac{N}{\Delta x}}$. The GPR is retrained once more using $\mathcal{D}^{(n)}$, this time optimising $\bm{\theta}$ by maximising the log marginal likelihood according to equation \eqref{eq:optim_theta}, yielding $\hat{\bm{\theta}}$ and $\hat{\sigma}^2_y$. This maximisation scheme chooses kernel hyperparameters and noise variance which make $\mathcal{D}^{(n)}$ seem probable. This optimisation is only carried out once the final observation set $\mathcal{D}^{(n)}$ has enough pixels to reconstruct the EoI. This optimisation drives $\sigma_y^2$ toward 0 because there are a sufficient enough edge pixels in $\mathcal{D}^{(n)}$ to interpolate through the observations exactly. Note via equation \eqref{eq:gp_updated_mean} that the smaller $\sigma^2_y$ is to 0, the closer the posterior predictive mean is to all observations. It is this optimisation scheme which converges the final PPD to the EoI, using the optimised posterior predictive mean,
\begin{equation}\label{eq:optimal_predict_mean}
    \hat{\bm{f}}^{(n+1)} \hspace{3pt}\big|\hspace{3pt}  \mathcal{D}^{(n)}, \hat{\bm{\theta}} = K_{*n}\left[K_{nn}+ \hat{\sigma}^2_y\mathbb{I}\right]^{-1}\bm{y}^{(n)},
\end{equation}
as the proposed edge. This, as well as a 95\% credible band, evaluated at $X^*$, is outputted to the user. Algorithm \ref{alg:gpet} outlines the pseudocode of the proposed edge tracing algorithm.

\begin{algorithm}[t!]\footnotesize
    \caption{Edge Tracing using Gaussian Process Regression}
    \textbf{Inputs}: Kernel $k$ and its hyperparameters $\bm{\theta}$, including observation noise variance $\sigma_y^2$, image gradient $G$, edge endpoints $\mathcal{D}^{(0)}$, number of posterior curves $L$ and proportion to keep as optimal $\epsilon$, score threshold $T$ and sub-interval length $\Delta x$. \\
    \textbf{Output}: Optimal posterior predictive mean with a 95\% credible band, evaluated at $X^*$.
    \begin{algorithmic}[1]
        \State Induce prior predictive distribution, $f^{(0)}(x) \sim \mathcal{GP}\big(0, k(x, x' ; \bm{\theta})\big)$;
        \State Fit edge endpoints in $\mathcal{D}^{(0)}$ to the GPR and induce the initial PPD, $\bm{f}^{(1)} | \mathcal{D}^{(0)}$, corrupting edge endpoints with observation noise $\sigma^2_y$;
        \While{$|\mathcal{D}^{(n)}| \neq \ceil{\nicefrac{N}{\Delta x}}$}
            \State \multiline{Draw $L$ posterior curves from the current PPD;}
            \State \multiline{Compute cost of each posterior curve and select $\epsilon L$ curves which have the least cost;}
            \State \multiline{Estimate frequency density function $\phi^{(n)}$ using optimal posterior curves, weighted by their curve score;}
            \State \multiline{Score and threshold pixel coordinates, yielding $P^{(n)}_T$;}
            \State \multiline{Bin and use non-max suppression to obtain $P^{(n)}_A$;}
            \State \multiline{Combine $\mathcal{D}^{(n-1)}$ and $P^{(n)}_A$ to obtain $\mathcal{D}^{(n)}$;}
            \State {While $|\mathcal{D}^{(n)}| \leq |\mathcal{D}^{(n-1)}|$, reduce $T$ and repeat steps 7--9;}
            \State \multiline{Retrain the GPR to induce an updated PPD, $\bm{f}^{(n+1)} | \mathcal{D}^{(n)}$, corrupting $\mathcal{D}^{(n)}$ with observation noise $\sigma^2_y$.}
        \EndWhile
    \State \multiline{Optimise hyperparameters $\bm{\theta}$ and retrain GPR using final set of observations $\mathcal{D}^{(n)}$ with optimised hyperparameters $\hat{\bm{\theta}}$;}
    \State \textbf{return} optimised posterior predictive mean, $\hat{\bm{f}}^{(n+1)} | \mathcal{D}^{(n)}, \hat{\bm{\theta}}$ evaluated at $X^*$, alongside a 95\% credible band.
    \end{algorithmic}
    \label{alg:gpet}
\end{algorithm}

\section{Applications}\label{apps}
In order to illustrate the effectiveness of the proposed methodology, we will compare its performance with two commonly used segmentation algorithms in medical and satellite image applications. These two segmentation algorithms are Dijkstra's shortest path algorithm \cite{dijkstra1959note} and the active contour, energy minimising snake algorithm \cite{kass1988snakes}, both of which are available from the scikit-image Python library \cite{Walt2014skimage}. Table \ref{tab:results_table} summarises the performance between the proposed methodology and the two popular edge tracing techniques using the Jaccard score \cite{jaccard1912distribution} for comparing intersection-over-union of two binary segmentation masks. Execution time is also measured but with no expectation that the proposed algorithm will be faster given the scikit-image implementation of Dijkstra's algorithm is heavily optimised using Cython \cite{Walt2014skimage}, unlike the proposed and active contour approaches. Due to the Mat\'{e}rn kernel having better capacity for modelling real-life processes, all edges traced using our approach in this section utilise the Mat\'{e}rn kernel with $\nu = 2.5$. Note, for ease of reading, the 95\% credible intervals are not shown in the applications of this section.

\subsection{Choroid Segmentation in OCT Imaging}\label{subsec:OCT}
Optical coherence tomography (OCT) imaging acquires micro-resolution, three-dimensional maps of biological tissue using low-coherence light. This technique is primarily used in ophthalmology to image the retinal and choroidal structures at the back of the eye. They are typically used to help clinicians deduce pathologies which affect or can be observed in the eye such as  glaucoma \cite{lederer2003analysis}, chronic kidney disease \cite{balmforth2016chorioretinal} and cardiovascular disease \cite{farrah2020eye}. Because the layer boundaries in the retina and choroid are predominantly horizontal, this is a perfect application for the proposed methodology. The lower choroid boundary, the choroid--scleral interface, is particularly challenging to trace because of vessel shadowing and speckle noise. The latter is the result of the signal weakening the further it penetrates through the eye.

\begin{figure}[!t]
    \centering
    \setlength{\unitlength}{1cm}
    \begin{picture}(20, 7.65)
        \put(0.0,0.0){\includegraphics[width=\textwidth]{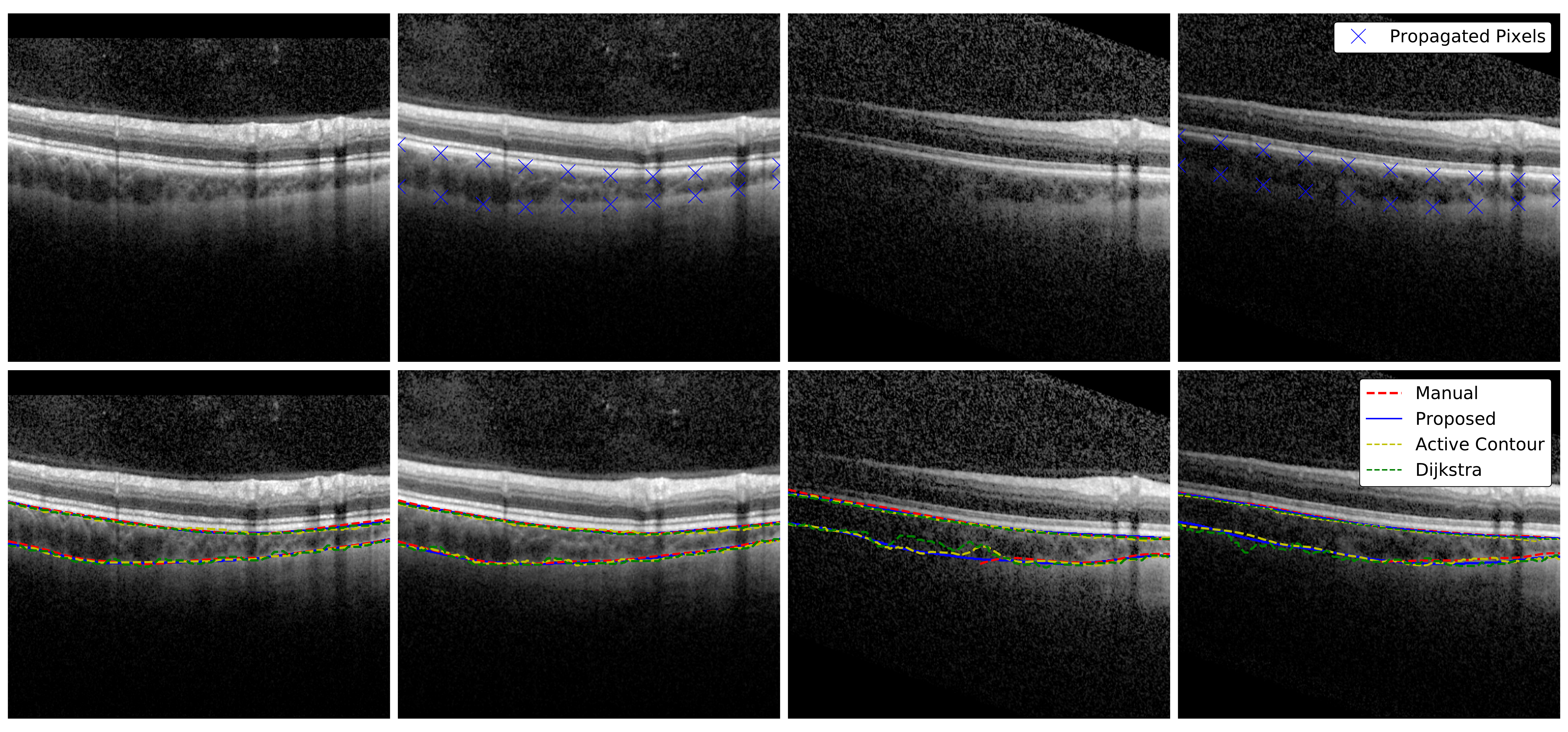}}
        \put(0.175, 7.2){\textbf{\color{white}(a)}}
        \put(4.3, 7.2){\textbf{\color{white}(b)}}
        \put(8.375, 7.2){\textbf{\color{white}(c)}}
        \put(12.475, 7.2){\textbf{\color{white}(d)}}
        \put(0.175, 3.45){\textbf{\color{white}(e)}}
        \put(4.3, 3.45){\textbf{\color{white}(f)}}
        \put(8.375, 3.45){\textbf{\color{white}(g)}}
        \put(12.475, 3.45){\textbf{\color{white}(h)}}
    \end{picture}
    \caption{(a--d) Four cross-sectional OCT images. (b, d) Points propagated from traces in (e, g) in blue (more on this in section \ref{subsec:propagate}). (e--h) Choroid segmentation of the three automated edge tracing techniques with manual labelling.}
    \label{fig:OCT_traces}
\end{figure}
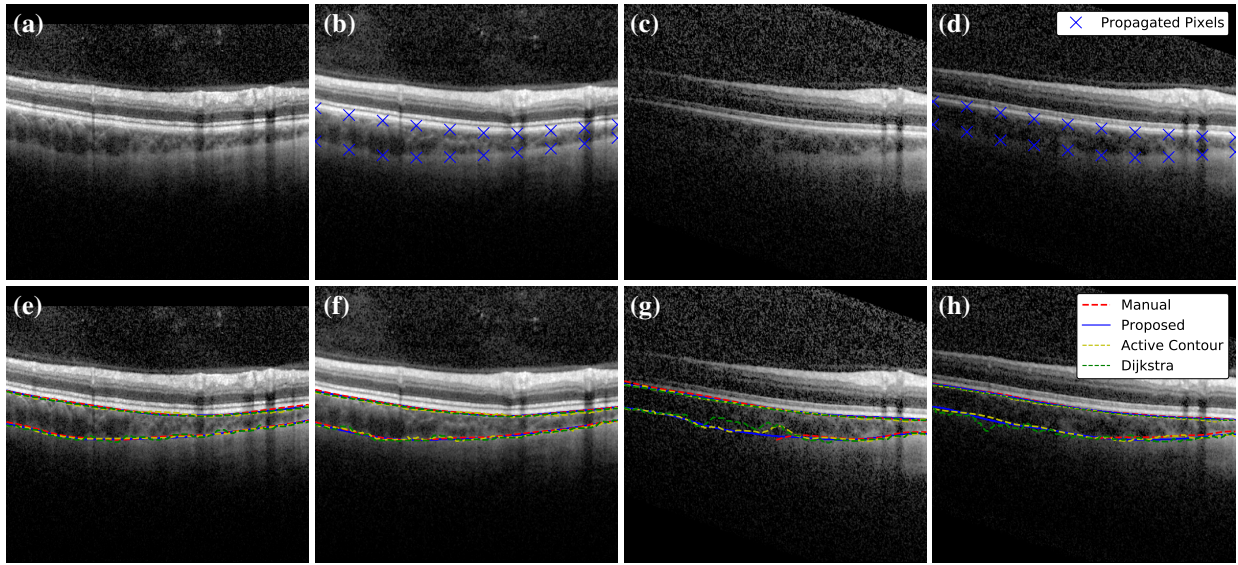
Figure \ref{fig:OCT_traces} shows four scans from two exemplar OCT image sequences. Subfigures \ref{fig:OCT_traces}(a, b) and \ref{fig:OCT_traces}(c, d) are consecutive pairs of the same sequence. The bottom row shows the result of segmenting the choroidal region using the three automated approaches with manual segmentations provided by a trained, clinical ophthalmologist. The two OCT image sequences highlight the extremes in image quality of OCT cross-sectional scans. While subfigures \ref{fig:OCT_traces}(e, f) show a clearer choroid region successfully segmented by all three approaches, subfigures \ref{fig:OCT_traces}(g, h) illustrate the proposed algorithms superiority in dealing with low signal strength. This is due to utilising its Gaussian process back-end to interpolate through missing and noisy data. Note that the manual segmentation was unable to complete the lower choroid-scleral boundary in subfigures \ref{fig:OCT_traces}(g, h).

\subsection{Optic Disc Segmentation in Fundus Imaging} \label{subsec:fundus}
The optic disc is an approximately ellipsoidal retinal structure defining the outer rim of the optic nerve head (ONH). This is the the part of the eye which allows retina vessels to carry visual information to the brain. The ONH is usually shown to be a brighter region on the fundus than anywhere else. Accurate segmentation of the optic disc along the optic nerve head's circumference is the first step toward diagnosis of glaucoma \cite{maccormick2019accurate}. Since the proposed methodology must enforce injectivity when modelling an EoI, the optic disc cannot immediately be traced. However, since it is a closed curve, we can utilise polar coordinate transformation so that the EoI becomes injective.  

\begin{figure*}[!t]
    \centering
    \setlength{\unitlength}{1cm}
    \begin{picture}(20, 8.75)
        \put(0.0,0.0){\includegraphics[width=\textwidth]{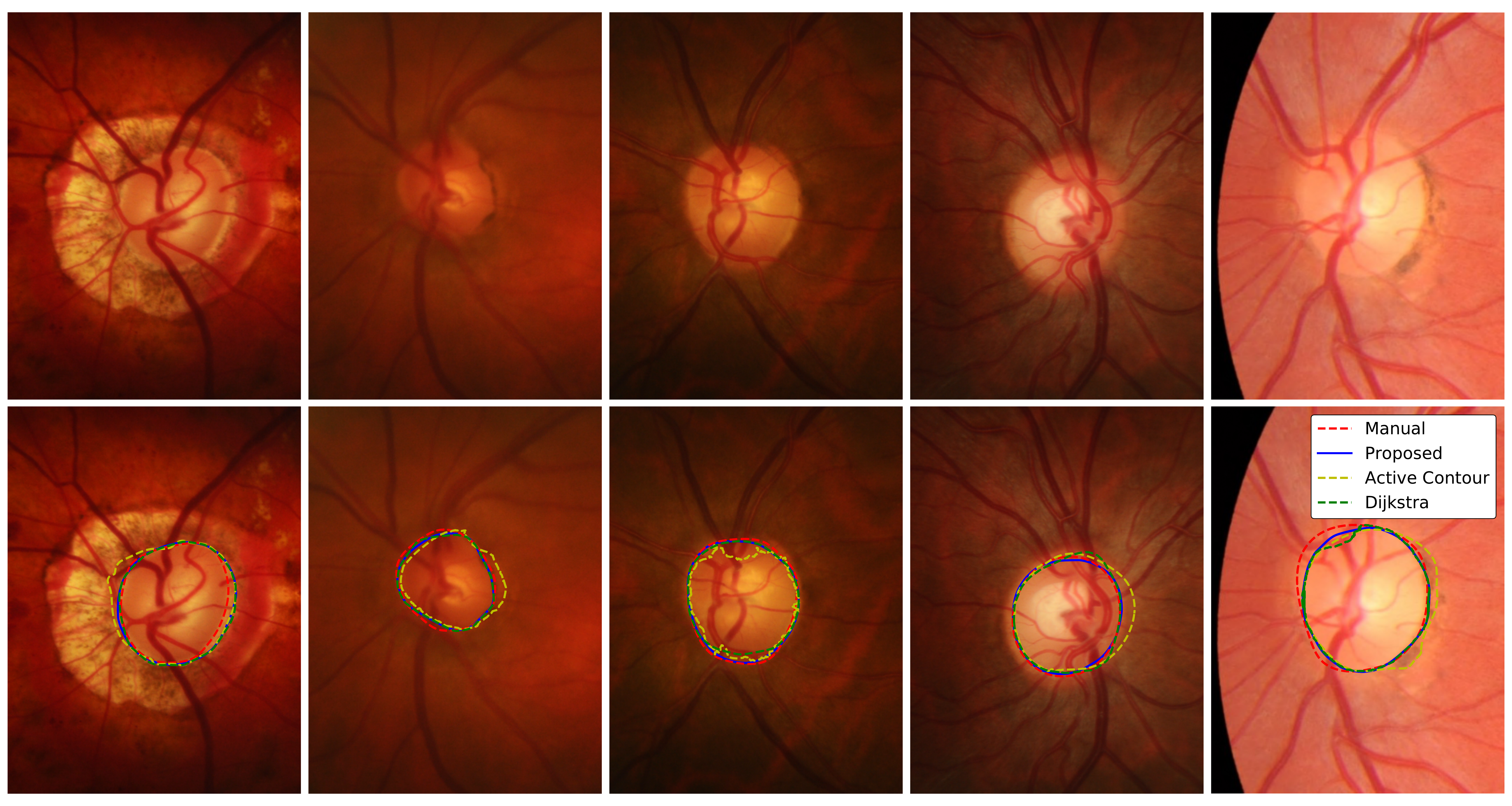}}
        \put(0.175, 8.3){\textbf{\color{white}(a)}}
        \put(3.45, 8.3){\textbf{\color{white}(b)}}
        \put(6.75, 8.3){\textbf{\color{white}(c)}}
        \put(10.05, 8.3){\textbf{\color{white}(d)}}
        \put(13.3, 8.3){\textbf{\color{white}(e)}}
        \put(0.175, 4){\textbf{\color{white}(f)}}
        \put(3.45, 4){\textbf{\color{white}(g)}}
        \put(6.75, 4){\textbf{\color{white}(h)}}
        \put(10.05, 4){\textbf{\color{white}(i)}}
        \put(13.3, 4){\textbf{\color{white}(j)}}
    \end{picture}
    \caption{(a--e) Five fundus images showing the optic disc and retinal vessels. The first two are glaucomatous and the remaining three are healthy. (f--j) Optic disc segmentations of the three automated edge tracing approaches with manual labelling.}
    \label{fig:fundus_traces}
\end{figure*}
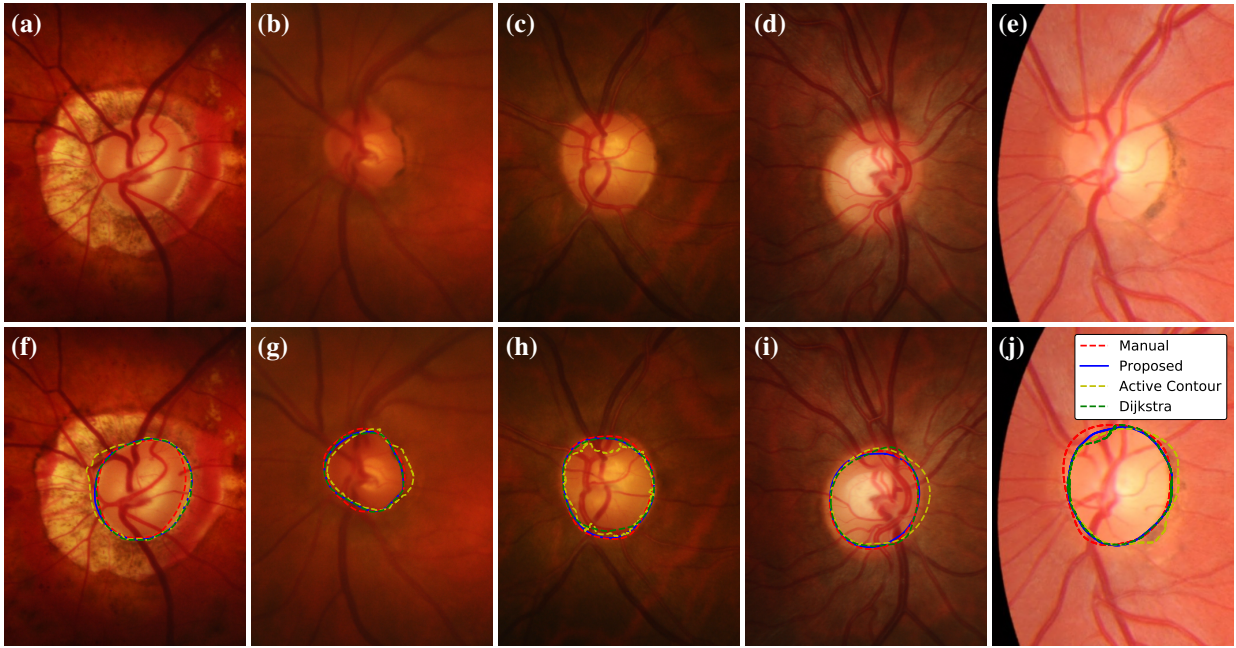

Figure \ref{fig:fundus_traces} shows five fundus scans acquired from the RIM-ONE open database of healthy and glaucomatous images \cite{fumero2011rim} and the scikit-image Python data module \cite{Walt2014skimage}. Here, each EoI is occluded at several points as retinal vessels enter the disc. The proposed methodology consistently interpolates around the retinal vessels while the other automated approaches fall victim either to the inner retinal vessels or extraneous tissue surrounding the disc, as seen in subfigures \ref{fig:fundus_traces}(h, j) or \ref{fig:fundus_traces}(f, g, i), respectively. In particular, Dijkstra's algorithm fails due to its greedy one-shot approach, which always prefers the path of maximal image gradient response without taking into account the holistic geometric properties of the EoI.

\subsection{Satellite Image Segmentation}\label{subsec:satellite}
\begin{figure}[!t]
    \centering
    \setlength{\unitlength}{1cm}
    \begin{picture}(20, 5)
        \put(0.0,0.0){\includegraphics[width=\textwidth]{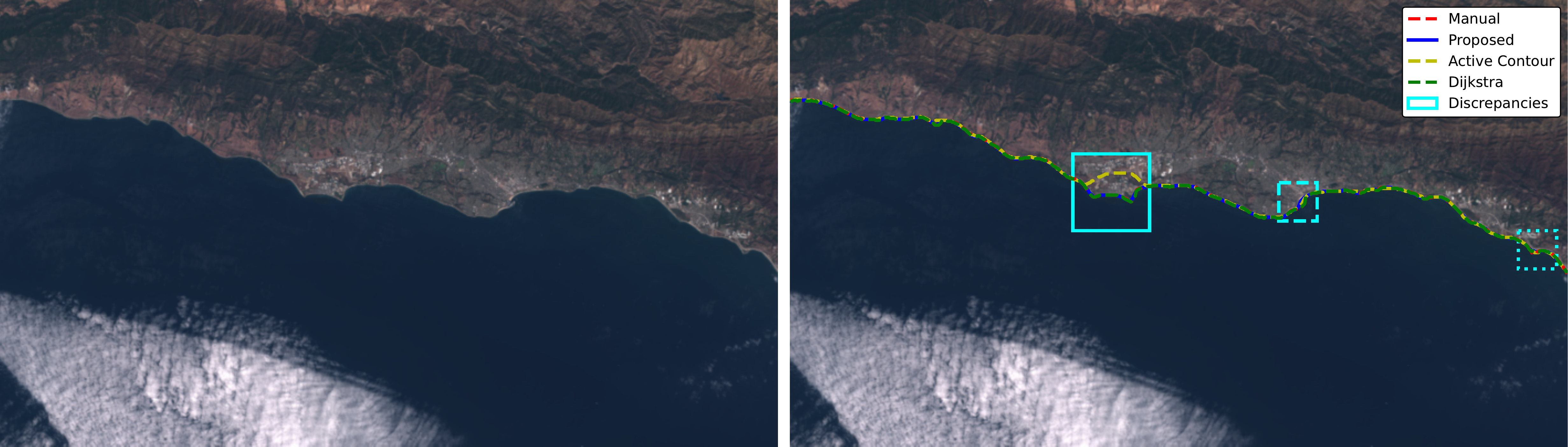}}
        \put(0.1, 4.35){\textbf{\color{white}(a)}}
        \put(8.45, 4.35){\textbf{\color{white}(b)}}
    \end{picture}
    \caption{(a) Satellite image showing the coast of California. (b) Coast segmentations of the three automated edge tracing techniques with manual labelling. Coloured squares represent discrepancies between traces.}
    \label{fig:sat_traces}
\end{figure}

Another important application of image processing is satellite imaging. This can help locate and map land and water disasters such as landslides \cite{yu2020landslide} or mass oil spills \cite{yekeen2020novel}. Figure \ref{fig:sat_traces} shows a satellite image of part of the coast of California, acquired from the European Space Agency's Copernicus program \cite{copernicusSatelliteHub}. Traces from all three automated segmentation approaches and manual labelling are shown in the bottom subfigure. The satellite image was chosen because the EoI is predominantly injective and geometrically complex, which can pose problems of short-circuiting. Short-circuiting is where a segmentation algorithm prefers a cheaper or smoother path to follow rather than the true edge, which may contain sharp corners for example. 

The proposed methodology and Dijkstra's algorithm trace the coast successfully with minor issues of short-circuiting toward the right-hand side of the image, shown in the blue dotted square. Their main region of disagreement can be seen in the blue dashed square, where the coast becomes temporarily multi-valued which Dijkstra's algorithm can successfully trace, unlike the proposed methodology which short-circuits. However, the active contour model performs the most poorly, particularly in the region highlighted by the blue filled square. One of the difficulties of the active contour model is in knowing how to tune its parameters correctly, particularly the initial contour which is crucial and can be difficult to define a priori. 

\subsection{Image Sequence Segmentation}\label{subsec:propagate}
The GPR at the beginning of the algorithm need not only be initialised with an estimate for the edge endpoints. Instead, the algorithm can be initialised with an evenly spaced set of pixels spanning the EoI. A use case for this feature is taking advantage of domain knowledge in image sequences, such as frames of a video or tomographic data. By tracing an EoI in one frame of the sequence, we can propagate a subset of these edge pixels into the model when tracing the same edge in consecutive images. This is under the assumption the edge changes incrementally, maintaining most of the same geometric and structural properties, between consecutive images. This allows the algorithm to start with a PPD which samples curves in the vicinity of the new edge, reducing the algorithms exploration space, quickening convergence along the new edge compared with initialising using only the edge endpoints.

To refer back to subsection \ref{subsec:OCT}, subfigures \ref{fig:OCT_traces}(b, d) show two cross-sectional scans next in the sequence from scans in \ref{fig:OCT_traces}(a, c). Because OCT data represents a volumetric scan of biological tissue, there is necessarily some consistency and similarity in the geometric properties of the choroid and retinal interfaces between consecutive scans. Subfigures \ref{fig:OCT_traces}(b, d) show blue, crossed points extracted from the traces of the proposed methodology in subfigures \ref{fig:OCT_traces}(e, g), respectively. Using these pixels as initialisation, the proposed methodology successfully segments the choroid in subfigures \ref{fig:OCT_traces}(f, h). Table \ref{tab:results_table} summarises the performance across all three automated segmentation approaches compared to manual segmentations.

\begin{table}[!t]
    \centering
    \begin{tabular}{|c|c|l|l|l|}
        \hline
        \multicolumn{2}{|c|}{\multirow{2}{*}{\textbf{Application}}} & \multicolumn{3}{c|}{\textbf{Method}} \\ \cline{3-5} 
        \multicolumn{2}{|c|}{} & \multicolumn{1}{c|}{Proposed} & \multicolumn{1}{c|}{\begin{tabular}[c]{@{}c@{}}Active Contour \\ \cite{kass1988snakes}\end{tabular}} & \multicolumn{1}{c|}{\begin{tabular}[c]{@{}c@{}}Dijkstra \\ \cite{dijkstra1959note}\end{tabular}} \\ \hline
        
        \multirow{2}{*}{\begin{tabular}[c]{@{}c@{}}\textbf{OCT} \\ (Fig. \ref{fig:OCT_traces})\end{tabular}} & $J$ (\%) & $\mathbf{91.3}$ ($\pm$ 6.0) & 86.5 ($\pm$ 5.5) & 85.2 ($\pm$ 5.6) \\ \cline{2-5} 
         & Time (s) & 8.2 ($\pm$ 11.5) & 0.9 ($\pm$ 1.5) & $\mathbf{0.023}$ ($\pm$ 0.01) \\ \hline
         
        \multirow{2}{*}{\begin{tabular}[c]{@{}c@{}}\textbf{Fundus} \\ (Fig. \ref{fig:fundus_traces})\end{tabular}} & $J$ (\%) & $\mathbf{91.8}$ ($\pm$ 4.1) & 82.8 ($\pm$ 2.4) & 89.9 ($\pm$ 3.0) \\ \cline{2-5} 
         & Time (s) & 28.1 ($\pm$ 11.7) & 32.3 ($\pm$ 37.0) & $\mathbf{0.109}$ ($\pm$ 0.1) \\ \hline
         
        \multirow{2}{*}{\begin{tabular}[c]{@{}c@{}}\textbf{Satellite} \\ (Fig. \ref{fig:sat_traces})\end{tabular}} & $J$ (\%) & 99.7 & 98.1 & \textbf{99.9} \\ \cline{2-5} 
         & Time (s) & 100.5 & 21.0 & \textbf{0.05} \\ \hline 
         
        \multirow{2}{*}{\begin{tabular}[c]{@{}c@{}}\textbf{Sinusoid} \\ (Fig. \ref{fig:kde_plot})\end{tabular}} & $J$ (\%) & \textbf{99.6} & 93.1 & 77.0 \\ \cline{2-5} 
         & Time (s) & 38.8 & 13.5 & \textbf{0.1} \\ \hline
    \end{tabular}
    \caption{Execution time and accuracy of all three automated edge-based segmentation techniques. Accuracy was measured using the Jaccard score, $J$.}
    \label{tab:results_table}
\end{table}

\section{Discussion}\label{discussion}
The results in section \ref{apps} highlight the benefits of the proposed methodology. Primarily, the novel technique is able to successfully trace discontinuous edges by interpolating between high gradient responses in a way that ensures a coherent overall trace for a given edge. This is, in part, due to converting the discrete optimisation problem to a continuous one using Gaussian process regression. This is advantageous for tracing edges that are relatively smooth, as having knowledge of some pixel coordinates in the image allows the GPR to sample curves which make appropriate assumptions on the path of the edge in neighbourhoods of such coordinates. Moreover, it can quantify uncertainty in regions of discontinuity and efficiently search it for adequate pixel coordinates which conform to the properties of the user-specified kernel, and also to the properties of the EoI when pixel coordinates surrounding these discontinuities are known. This allows the model to intelligently bridge gaps in otherwise discontinuous edges which may be degraded by noise or occluded by other artefacts. 

Our novel approach uses the image gradient to provide local edge information of potential edge pixels and combines this with global, structural information from the frequency distribution of optimal posterior curves. This is a direct consequence of modelling the EoI using Gaussian process regression and utilising its predictive variance. In each new iteration, the most optimal posterior curves are used to recalculate the frequency distribution. This frequency distribution shows higher density in regions where the most optimal posterior curves have visited. This allows the accept-discard scheme in each iteration to score previously and newly accepted pixel coordinates, dynamically refining the observation set of pixel coordinates depending on this frequency distribution. The balance between the image gradient and frequency distribution in equation \eqref{eq:score_eqn} is key to deciding on the next set of pixels to fit to the GPR. Placing too much weighting on either image gradient or frequency distribution can have undesired effects. In the former, the model could be perturbed into accepting high-gradient, noisy non-edge pixels, while in the latter, local information is ignored in favour of selecting pixels which have been visited most, regardless of image gradient response. 

To allow the GPR to continue exploring the space throughout the fitting procedure, the kernel hyperparameters and observation noise are not optimised in each iteration, contrary to typical application of Gaussian process regression. In our recursive Bayesian scheme, it is not the hyperparameters we wish to update but the PPD, utilising its corresponding predictive variance and posterior curves given a refined, but incomplete, observation set. Performing hyperparameter optimisation with an incomplete observation set, which necessarily have regions with missing data, will select kernel hyperparameters according to the current observations fitted to the GPR. Only posterior curves which fit the incomplete observation set will be sampled from then on, reducing the predictive variance in regions with missing data. This will limit the flexibility the PPD has in imputing the empty regions in later iterations. A disadvantage here is that an incorrect choice of kernel and hyperparameters could lead to a sub-optimal edge prediction due to the incorrect function properties encoded at the start of the algorithm. 

Throughout we have assumed that the edge's dominant direction is along the image width. We have already shown the use of polar image transformation to deal with simple, closed and elliptical edges. For edges whose dominant direction is not horizontal, a simple solution relies on being able to draw a line between the edge endpoints and checking that the edge does not appear multi-valued along that axis. Provided this is the case, then simple image rotation will convert the edge's dominant direction to be horizontal. Moreover, edges whose dominant direction changes in different regions of the image can be split into sub-edges and each can be traced individually, ensuring each sub-edge is injective along the image width. However, this introduces the extra computational overhead of estimating the edge endpoints for each sub-edge. We intend to address the issue of more complex and multi-valued edges in future work on this problem.

\subsection{Sensitivity Analysis}\label{subsec:sens_anal} 
\begin{figure*}[!t]
    \centering
    \includegraphics[width=\textwidth]{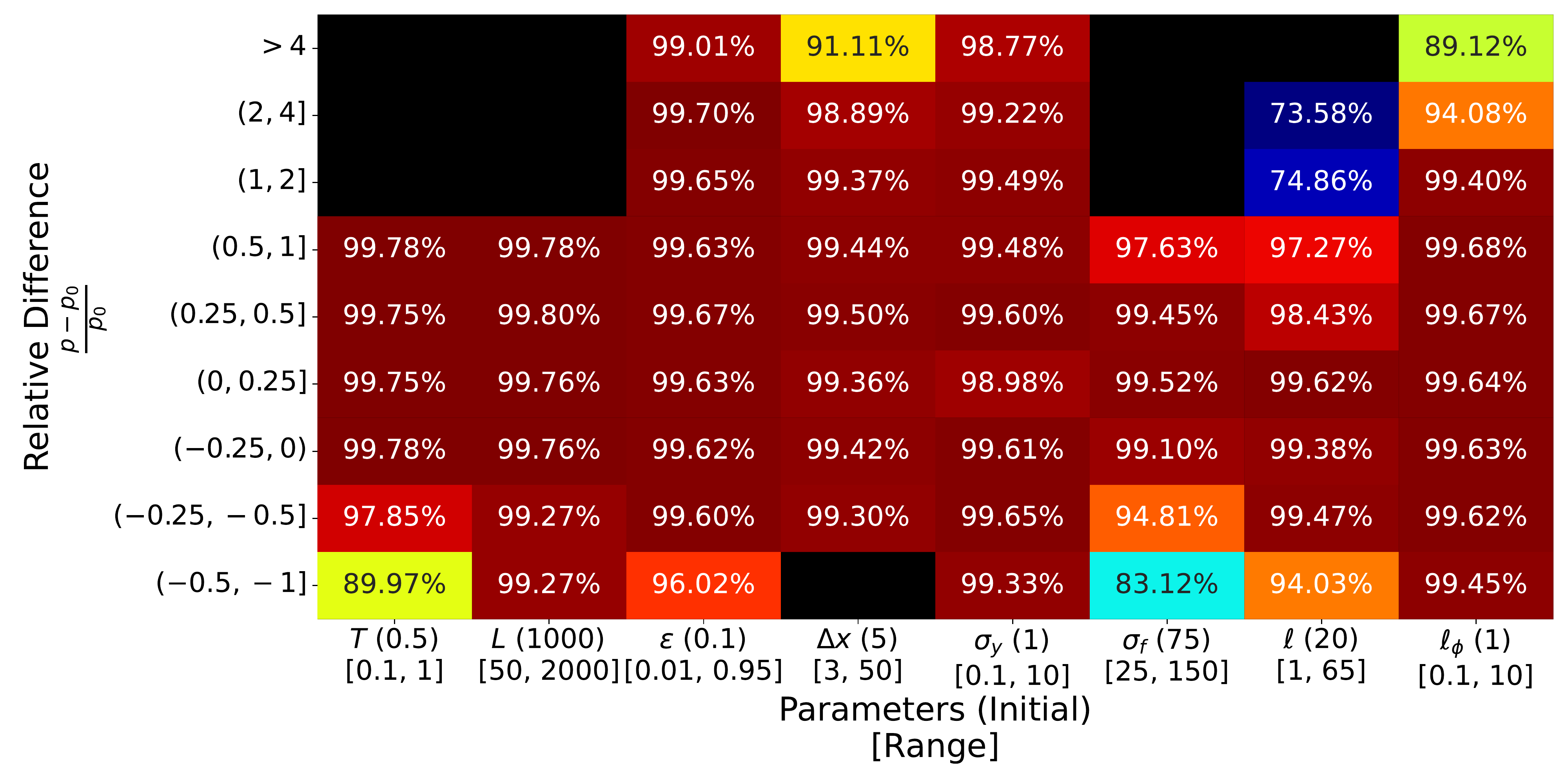}
    \caption{Sensitivity analysis using test case shown in figure \ref{fig:kde_plot}. Heatmap annotated with average Jaccard scores for individual parameter perturbations, $p = p_0 + \delta$, grouped according to relative change in parameter value from an initial choice, $p_0$.}
    \label{fig:sensitivity_analysis}
\end{figure*}

In order to assess the robustness of the proposed methodology and its user-defined parameters, a comprehensive sensitivity analysis was conducted. This analysis used the test case in figure \ref{fig:kde_plot} and the Jaccard score to measure how well the proposed methodology deals with perturbations in the user-defined input parameters. An edge prediction using an initial, pre-defined choice of parameters was compared with edge predictions resulting from perturbations of individual parameters, keeping all others fixed. Figure \ref{fig:sensitivity_analysis} shows the results for assessing the robustness of eight model hyperparameters; score threshold $T$, number of posterior curves $L$, proportion of curves to keep $\epsilon$, sub-interval length $\Delta x$, noise variance $\sigma^2_y$, signal variance $\sigma_f^2$, lengthscale $\ell$ and frequency distribution lengthscale $\ell_\phi$. For each of the eight hyperparameters, 100 values were chosen uniformly at random within a range of relative differences of over 5 times their initial value, resulting in 800 different experiments. The relative differences were chosen based on realistic perturbations from pre-defined, initial values. This meant excluding relative differences which were impossible, such as $T > 1$, or unnecessary, such as $L > 2000$ as it clearly adds no value or $\ell > 65$ as there are already diminishing returns in Jaccard score for $\ell > 50$ in figure \ref{fig:sensitivity_analysis}.

The sensitivity analysis provides an empirical guide for selecting parameters. The most robust parameters are the number of posterior curves $L$, keep ratio $\epsilon$ and noise variance $\sigma^2_y$, as there is negligible impact on edge prediction for almost all perturbations. $L$ should not be arbitrarily large as it has computational overhead of GPR sampling and costing of posterior curves. A value of $L$ between 100 and 1000 will suffice. $\epsilon$ has little impact on edge prediction due to the weighting scheme in forming the frequency density function. A very small value is undesirable as the model is forced to rely on a small subset of posterior curves to choose pixels with. $\sigma^2_y$, encoding the confidence in the models' pixels, $\mathcal{D}^{(n)}$, should be selected between 0 and 2. The lower the value of $\sigma^2_y$ implies a lower predictive variance in regions near observations in the PPD. Too large a value implies the model lacks confidence during pixel selection which can reduce accuracy.

Other parameters which are slightly more sensitive are the score threshold $T$, sub-interval length $\Delta x$ and lengthscale of the kernel used in constructing the frequency distribution $\ell_\phi$. A small score threshold or large sub-interval length quickens convergence since the margin for pixel acceptance is low or the number of observations in $\mathcal{D}^{(n)}$ need not be large, respectively. However, this comes at a cost of accuracy. Because the score threshold is adaptive between iterations, setting $T = 1$ will initialise the search by only accepting pixels with a high gradient response and frequency density. $\Delta x$ defines how many pixels are used to reconstruct the EoI and the choice of $\Delta x$ depends on its complexity. For practical purposes, $\Delta x = 5$ can reliably reconstruct edges of varying complexity. $\ell_\phi$ was chosen to mimic the lengthscale of a pixel, and for values $\ell_\phi \geq 3$ the frequency density loses precision in where density is contributed. This causes density to spread to non-edge regions of the image leading to undesired effects, such as short-circuiting or acceptance of noisy, non-edge pixels.

Based on the results in figure \ref{fig:sensitivity_analysis}, the signal variance $\sigma_f^2$ and kernel lengthscale $\ell$ are the most sensitive user-defined parameters. $\sigma_f^2$ dictates the predictive variance of the PPD. This value should be chosen large enough to contain the EoI in the 95\% credible interval of the initial PPD, as seen in figure \ref{fig:init_post_dist} using $\sigma^2_f = 75^2$. The larger the kernel lengthscale $\ell$ is, the less sinuous the posterior curves are. This prevents the model from reconstructing the complexity of the EoI, explaining why we see the Jaccard score reduce for $\ell > 40$. The kernel lengthscale should never be set higher than the minimum distance between turning points in the EoI. Moreover, a smaller lengthscale results in highly sinuous posterior curves which slows convergence and begins to affect accuracy.

\section{Conclusion}\label{concl}
In this paper, we have proposed a novel technique for edge tracing using Gaussian process regression. The algorithm follows a recursive Bayesian scheme, intelligently exploring the image for edge pixels using a well informed PPD. The algorithm combines local edge information and global structural information during the pixel scoring procedure, with the heuristics of the selection and convergence procedures based on Bayesian methods. Our algorithm is able to traverse and interpolate discontinuous edges through its ability to model uncertainty in gaps of the domain of the observation set, searching for the most sensible pixels to bridge said gap. This technique provides the user with a set of robust and simple parameters to define at initialisation, controlling aspects of convergence and model complexity.

Moreover, the algorithm can be applied to numerous different image problems, a number of which we have illustrated and we have made favourable comparisons with two commonly used edge tracing algorithms. At present, the novel technique is unable to trace edges which are multi-valued along the predominant edge direction since it assumes injectivity. Many edges of interest have this property, particularly in medical imaging applications where we expect this algorithm to play a useful role. The proposed methodology can also be used efficiently for edge tracing in image sequences by utilising its ability to propagate a priori information from the edge in a previous frame of the sequence to new frames, limiting the search region to the vicinity of the new edge and speeding convergence.

\section*{Acknowledgements}
This work was supported by the Medical Research Council [grant number MR/N013166/1], as part of the Precision Medicine DTP with the University of Edinburgh. The authors would like to acknowledge Dr. Ian MacCormick, a clinical ophthalmologist at the University of Edinburgh for providing the manual ocular segmentations shown in section \ref{apps}.

\bibliographystyle{unsrt}  
\bibliography{bibliography.bib}

\end{document}